\documentclass[runningheads]{llncs}
\usepackage[T1]{fontenc}
\usepackage{graphicx}
\usepackage{tcolorbox}
\usepackage{booktabs}
\definecolor{pastelblue}{rgb}{0.93,0.80,1.00}
\usepackage{adjustbox}
\usepackage{multirow}
\usepackage{arydshln}
\usepackage{orcidlink}
\usepackage{hyperref}
\hypersetup{colorlinks,allcolors=black}
\begin{document}
\title{Do LLMs Provide Consistent Answers to Health-Related Questions across Languages? 
}
%
%

\author{Ipek Baris Schlicht\inst{1,2}\orcidlink{0000-0002-5037-2203} \and
Zhixue Zhao \inst{3}\orcidlink{0000-0002-3060-269X} \and
Burcu Sayin \inst{4}\orcidlink{0000-0001-6804-127X} \and
Lucie Flek\inst{2,5}\orcidlink{0000-0002-5995-8454} \and
Paolo Rosso\inst{1,6}\orcidlink{0000-0002-8922-1242}}
\institute{Universitat Politècnica de València, Spain \\
\email{\{ibarsch@doctor,prosso@dsic\}.upv.es}
\and
Bonn-Aachen International Center for IT, Germany \\
\email{\{ibarissc,flek@bit\}.uni-bonn.de} \and
University of Sheffield, UK \\
\email{zhixue.zhao@sheffield.ac.uk} \and
University of Trento, Italy \\
\email{burcu.sayin@unitn.it} \and
Lamarr Institute for Machine Learning and Artificial Intelligence, Germany \and
ValgrAI Valencian Graduate School and Research Network of Artificial Intelligence, Spain\\
}

\authorrunning{Schlicht et al.}
\newif\ifproofread
\newcommand{\changemarker}[1]{%
\ifproofread
\textcolor{red}{#1}%
\else
#1%
\fi
}

\proofreadfalse

\maketitle              
\begin{abstract}

Equitable access to reliable health information is vital for public health, but the quality of online health resources varies by language, raising concerns about inconsistencies in Large Language Models (LLMs) for healthcare.
In this study, we examine the consistency of responses provided by LLMs to health-related questions across English, German, Turkish, and Chinese.
\changemarker{We} largely expand the HealthFC dataset by categorizing health-related questions by disease type and broadening its multilingual scope with Turkish and Chinese translations.
We reveal significant inconsistencies in responses that could spread healthcare misinformation. Our main contributions are 1) a multilingual health-related inquiry dataset with meta-information on disease categories, and 2) a novel prompt-based evaluation workflow that enables sub-dimensional comparisons between two languages through parsing. Our findings highlight key challenges in deploying LLM-based tools in multilingual contexts and emphasize the need for improved cross-lingual alignment to ensure accurate and equitable healthcare information.

\keywords{Multilingual Q\&A \and Healthcare Misinformation \and Consistency Evaluation \and Large Language Models}
\end{abstract}

\section{Introduction}
\changemarker{LLMs} have become powerful tools for NLP tasks and become widely accessible through chat interfaces to non-technical users. Their applications also extend to healthcare, where users consult LLM-based chat applications for health-related questions. \changemarker{LLMs} are primarily trained on data from online sources. However, the quality of health information online varies by language, reflecting differences in national healthcare policies and practices. For example, non-English online health content often has lower quality~\cite{weissenberger2004breast,lawrentschuk37health,davaris2017thoracic}. This could result in inconsistencies in health-related information provided by LLM-based applications across different languages.

\begin{figure*}[!b]
    \centering
    \begin{minipage}{0.55\textwidth}
        \centering
        \includegraphics[trim={0 0.5cm 0 .5cm},clip, width=\linewidth]{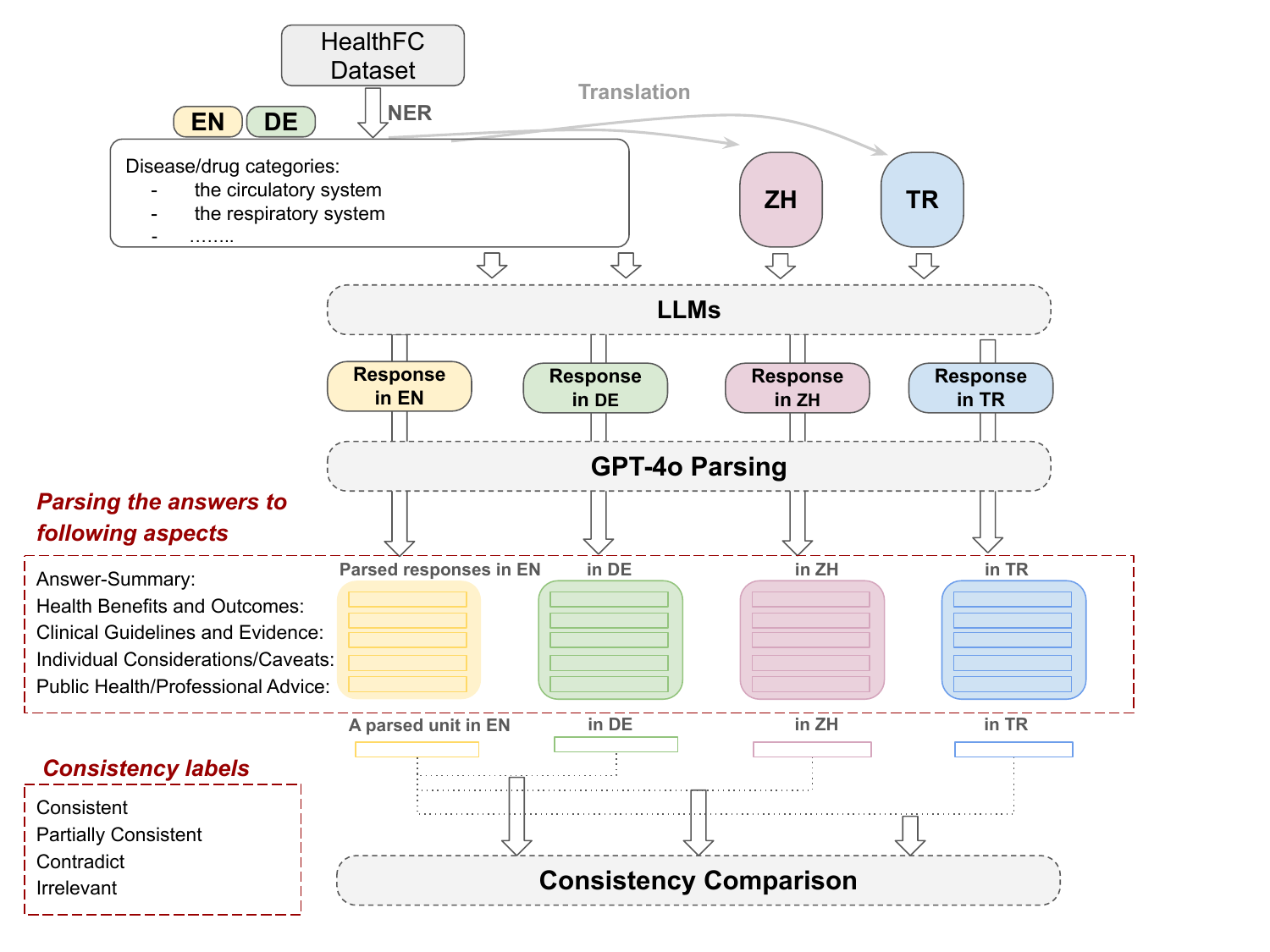}
        \caption{The proposed framework to automatically measure the consistency of health-related answers in EN, ZH, TR and DE.}
        \label{fig:first_figure}
    \end{minipage}%
    \hspace{0.04\textwidth} 
    \begin{minipage}{0.4\textwidth}
        \centering
        \scriptsize
        \caption{Distribution of 11 disease categories, each with at least 20 samples.}
      \adjustbox{width=\linewidth}{
        \begin{tabular}{lc}
            \toprule
            \textbf{Disease Category} & \textbf{Count} \\
            \midrule
            Symptoms, Signs \& Abnormal Findings  & 83 \\
            Neoplasms (Cancer)  & 61 \\
            Mental \& Behavioral Disorders & 46 \\
            Diseases of the circulatory system & 42 \\
            Diseases of the respiratory system & 36 \\
            Diseases of the nervous system & 31 \\
            Musculoskeletal \& Connective Tissue Diseases & 29 \\
            Etiology/Emergency use & 28 \\
            Injury, Poisoning \& External Causes & 26 \\
            Diseases of the digestive system & 22 \\
            Endocrine \& Metabolic Diseases & 22 \\
            \bottomrule
        \end{tabular}}
        \label{tab:disease_distribution}
    \end{minipage}
\end{figure*}

In this paper, we examine inconsistencies between responses provided in English and lower-resourced languages. These inconsistencies can introduce factual errors and may lead to biased medical guidance, potentially harming specific communities by perpetuating misinformation in one language while omitting it in another.
We particularly investigate the following research questions: \textbf{RQ1.} How do responses generated by an LLM differ when a question is asked in a non-English language compared to English?, and \textbf{RQ2.} Are there specific disease-related topics where LLM responses differ significantly across languages? To address these RQs, we first expand a health-related information-seeking dataset~\cite{vladika-etal-2024-healthfc} (originally available in English (EN) and German (DE)) to include Turkish (TR) and Chinese (ZH), categorizing responses by disease type. Next, we introduce a prompt-based evaluation framework, illustrated in Figure~\ref{fig:first_figure}, which segments the long-form responses of LLMs into informative parts to pinpoint areas with the highest occurrence of inconsistencies. \changemarker{Our} source code is available online\footnote{\url{http://bit.ly/4gQDJzT}}.

\section{Related Works}
Various studies have examined how prompt variations impact the factual consistency of LLMs in EN medical Q\&A. Zuccon and Koopman~\cite{koopman-zuccon-2023-dr} showed that the accuracy of ChatGPT can vary widely based on prompt phrasing, with supporting or contradicting evidence in the prompt potentially introducing biases that affect answer correctness. Similarly, Sayin et al.~\cite{sayin-etal-2024-llms}
explored the potential of LLMs to assist and correct physicians in medical decision-making.
They highlighted the role of prompt engineering in enhancing LLM interactions with medical experts, demonstrating its impact on LLMs’ ability to correct physician errors, explain medical reasoning, adapt to physician input, and improve overall performance.
Kaur et al.~\cite{kaur-etal-2024-evaluating} examined how presuppositions within prompts affect the factual accuracy and consistency of responses across LLMs such as ChatGPT and GPT-4. They showed that while LLMs rarely contradict established health facts, they often do not challenge false claims. Concerning multilingual Q\&A, Jin et al.~\cite{jin2024better} assessed ChatGPT’s response accuracy and consistency in multiple languages, translating non-EN responses into EN for evaluation. However, this method risks losing language-specific nuances due to its reliance on EN translations. While \changemarker{Jin et al.}~\cite{jin2024better} \changemarker{evaluated} consistency within individual languages, we \changemarker{assess} relative consistency across languages, focusing on EN (a high-resource language in multilingual LLM pretraining) compared to other languages.   
\section{Methodology}
To evaluate LLM performance across 
prompts in different languages, we first use Named Entity Recognition (NER) to categorize samples by disease. We then prompt LLMs in EN, DE, ZH, and TR on these categorized corpora.

\subsection{Information Seeker Queries}

We used the HealthFC dataset~\cite{vladika-etal-2024-healthfc}, available in DE and EN, containing 750 question-form claims along with corresponding evidence. HealthFC lacks specified disease categories for the claims, and some claims cover other health-related topics. To address this, we identified disease-related claims and enriched HealthFC with disease categories. Additionally, we expanded it by adding TR and ZH translations to support multilingual experiments.
\\
\noindent
\textbf{Disease Categorization.}
To identify disease-related claims in the dataset, we used HunFlair2~\cite{sanger2024hunflair2} from flairNLP~\cite{akbik2019flair} on EN claims. We selected samples containing disease entities, resulting in 508 samples. We used the ICD10 Coding~\cite{world1993icd} \changemarker{for the categorization}. We manually reviewed each entity, assigned the corresponding ICD10 codes, and then tagged them with their main disease category \changemarker{through the ICD10 look-up table~\cite{icd10}}. \changemarker{For example, the ICD10 code of skin cancer is C44.90 and hence its main disease category is Neoplasms according to the table}. After tagging, we kept only samples whose category occurred more than 20 times to maintain balance in disease categories (see Table~\ref{tab:disease_distribution}).
\\
\noindent
\textbf{Query Translations}
HealthFC includes samples in DE and EN. To expand it, we added TR and ZH samples by translating the EN samples with Google Translate. Native TR and ZH speakers from our team reviewed \changemarker{50 random} translations to ensure fluency and accuracy to preserve the original meaning. \changemarker{Specifically, they checked if the translated sentences and medical terms were correct and the translations retained the original meaning.} \changemarker{Despite noting areas for improvement}—such as using folk language instead of medical terms or adjusting culturally specific references (e.g., replacing 'pig meat' with 'sucuk' in TR), \changemarker{84\% of the TR translations and 98\% of the ZH translations were} satisfactory. 

\subsection{Prompt-based Evaluation Workflow}
\begin{figure*}[!b]
    \centering
    \begin{minipage}{0.5\textwidth}
        \centering
        \includegraphics[trim={0 3.6cm 0 0.4cm},clip, width=\linewidth]{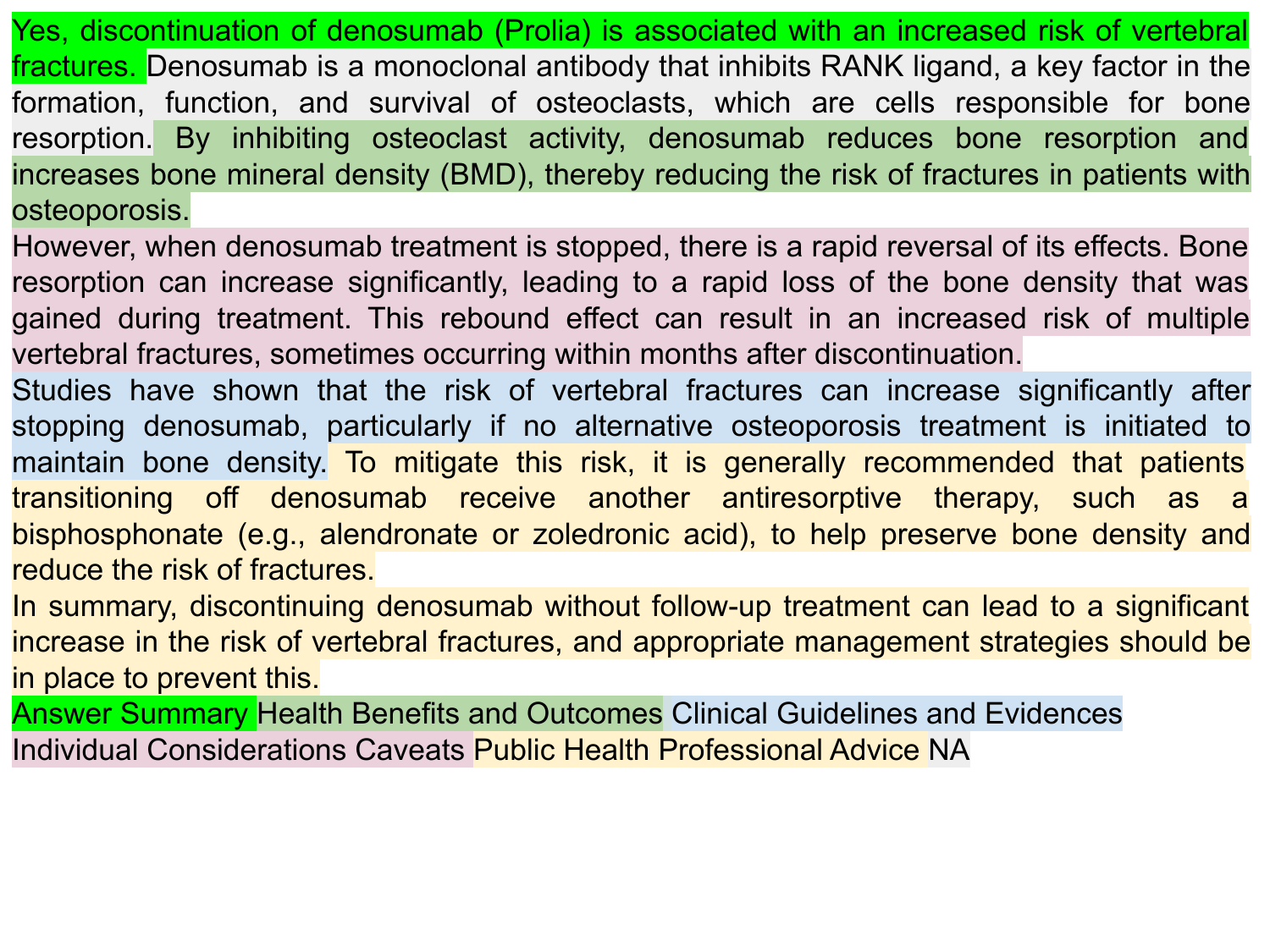}
    \end{minipage}%
    \begin{minipage}{0.5\textwidth}
    \includegraphics[trim={0 3.6cm 0 0.4cm},clip, width=\linewidth]{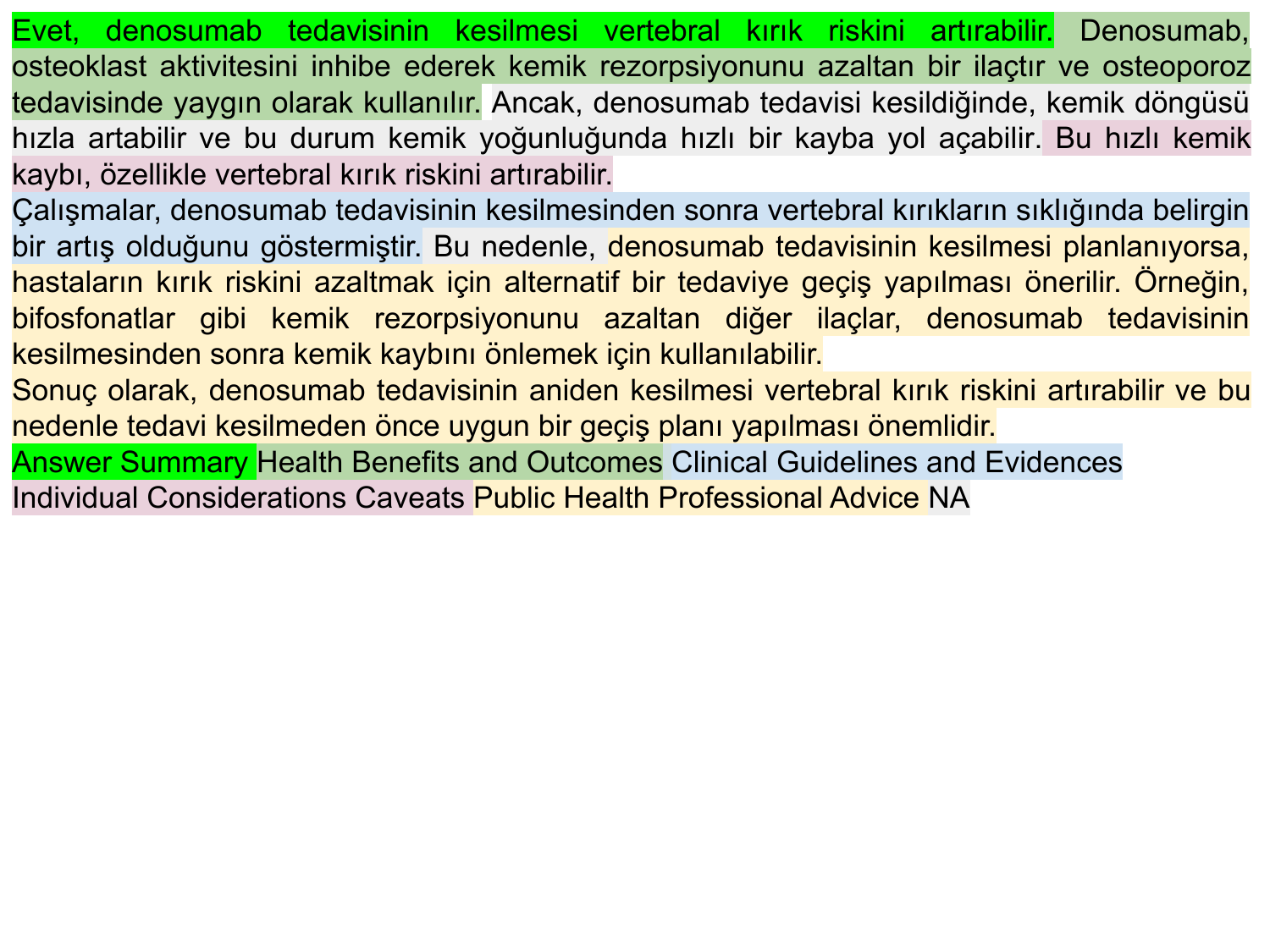}
    \end{minipage}
\caption{\changemarker{The answers to the question "Does discontinuation of denosumab increase the risk of vertebral fractures?": the EN answer on the left and the TR answer on the right. Both were parsed by GPT-4o using the proposed discourse ontology.}}
\label{fig:parsing}
\end{figure*}
LLMs generate complex long-form answers, unlike simpler formats like multiple-choice responses. Long answers may include not only direct responses but also background information, examples, limitations, and related content, making direct comparisons challenging. To address this, we developed a prompt-based evaluation workflow that automatically analyzes bilingual answer nuances by segmenting long-form responses into informative parts. This workflow includes a parsing prompt and a consistency-check prompt.
\\
\noindent
\textbf{Parsing Prompt.} This prompt parses the answer based on a discourse structure, inspired by prior work by Xu et al.~\cite{xu-etal-2022-answer} and adapted for our use case. The discourse ontology comprises the following elements: (i) \textit{Answer-Summary (AS):} The part of the answer addressing the question, excluding sentences elaborating on the summary or providing extra context, (ii) \textit{Health Benefits and Outcomes (HBO):} Describes the positive effects or results of a medical intervention or behavior, (iii) \textit{Clinical Guidelines and Evidence (CGE):} Refers to established guidelines or research that support the medical recommendations, (iv) \textit{Individual Considerations/Caveats (ICC):} Highlights individual variability and emphasizes the need for personalized advice, and (v) \textit{Public Health/Professional Advice (PHPA):} Emphasizes consulting healthcare professionals and following public health recommendations. \changemarker{A parsing example is given in Figure~\ref{fig:parsing}.}

\noindent\textbf{Consistency Comparison Prompt.} This prompt compares parsed EN answers with those in other languages, assessing whether they are consistent. Consistency is evaluated based on the following annotation schema: (i) \textit{Consistent:} The EN answer and the answer in the other language are fully consistent and semantically aligned, (ii) \textit{Partially Consistent:} The EN answer and the answer in the other language partially agree, overlap, or support each other, though with some irrelevant or contradictory content, (iii) \textit{Contradict:} Answers contradict each other, and (iv) \textit{Irrelevant:} Answers address different topics and are unrelated. Empty responses are also included in this irrelevant category.

\changemarker{We refined prompts iteratively and evaluated the final parsing prompt using a point-based system across five answers per language. Elements were scored: 2 points for correct classification, 1 point for partial parsing, with a maximum of 10. Average scores were 8.4 (EN, TR), 9.4 (ZH), and 7.2 (DE).} We used final prompts~\footnote{Final prompts can be accessed in our source code.}to parse and annotate the answers, and randomly selected 50 samples. \changemarker{The} team annotated the parsed answers in their native languages based on the inconsistency schema. \changemarker{We obtained} 250 annotations for DE and TR, and 240 annotations for ZH\footnote{Two examples were skipped by the annotator due to an error by Google Form}. We compared human annotations with LLM responses and observed disagreements between human and model annotations, especially within the inconsistent labels (partially inconsistent, irrelevant, contradiction). For automated evaluation, we consolidated these into a single `inconsistent' label, which resulted in higher Kappa Scores~\cite{mchugh2012interrater} between human and model evaluations across languages. The final Kappa scores showed substantial agreement for TR (K=0.66) and ZH (K=0.71), and moderate agreement for DE (K=0.50).

\begin{table}[b]
\caption{ \changemarker{\% of inconsistent responses by information units (AS: Answer Summary, HBO: Health Benefits and Outcomes, CGE: Clinical Guidelines and Evidence, ICC: Individual Considerations/Caveat, PHPA: Public Health/Professional
Advice.), with frequent variations when questions are asked in non-EN languages.}}
\adjustbox{width=\textwidth}{
\begin{tabular}{ccccccccccccc}
\toprule
& \multicolumn{4}{c}{\textbf{Chinese}} & \multicolumn{4}{c}{\textbf{Turkish}} & \multicolumn{4}{c}{\textbf{German}}   \\
& \textbf{ChatGPT} & \textbf{GPT4o} & \textbf{Llama3} & \textbf{CommandR+} & \textbf{ChatGPT} & \textbf{GPT4o} & \textbf{Llama3} & \textbf{CommandR+} & \textbf{ChatGPT} & \textbf{GPT4o} & \textbf{Llama3} & \textbf{CommandR+} \\
\midrule
\textbf{AS} & 19.95 & 24.18 & 31.22 & 37.56 & 17.61 & 28.64 & 31.46 & 35.21 & 14.55 & 23.24 & 25.59 & 27.23 \\
\midrule
\textbf{HBO}  & 36.15 & 40.61 & 57.51 & 61.74 & 44.37 & 46.48 & 61.74 & 62.91 & 33.10 & 30.75 &51.88 & 54.23 \\
\midrule
\textbf{CGE} & 47.65 & 42.02 & 69.01 & 71.83 & 53.05 & 49.06 & 77.93 & 73.00 & 40.61 & 38.03 & 73.94 & 65.26 \\
\midrule
\textbf{ICC} & 66.43 & 62.91 & 82.16 & 81.22 & 65.73 & 69.25 & 84.98 & 86.39 & 64.08 & 60.09 & 71.83 & 77.70 \\
\midrule
\textbf{PHPA} & 45.07 & 43.43 & 71.13 & 76.76 & 54.93 & 47.89 & 81.92 & 83.10 & 50.47 & 47.42 & 65.73 & 80.75  \\
\midrule
\textbf{Average} & 43.05 & 42.63 & 62.21 & 65.82 & 47.34 & 48.26 & 67.61 & 68.12 & 40.56 & 39.91 & 57.79 & 61.03 \\
\bottomrule

\end{tabular}}
    \label{tab:res_entailment}
\end{table}

\section{Experiments and Results}
We included four general-purpose, multilingual LLMs in the experiments: the closed-source models \changemarker{ChatGPT-4o, continuously updated model~\cite{openai-documentation}, and GPT4-o} from OpenAI~\cite{achiam2023gpt}, and the open-source models \changemarker{Llama3}-70B-Instruction~\cite{dubey2024llama} from Meta and CommandR+ from Cohere~\cite{cohere}. To generate answers, we set the LLMs' temperature to \changemarker{0} and the response context limit to 2048 tokens. Prompts were given in EN, as the models tend to follow task descriptions more effectively in EN than in other languages~\cite{lupo2023use,10.1145/3589335.3651902}, and EN is more cost-effective due to the larger token size required by some languages. No system prompts were used, except for Llama and CommandR+ when generating responses in TR and ZH, as these models default to EN. We accessed the closed models via the official OpenAI API. We used the Huggingface Inference Endpoint for Llama3 and the Transformers library~\cite{wolf2020transformers} for CommandR+. Additionally, GPT-4o was used for parsing and comparing responses, chosen for its strong performance on various NLP tasks and its support for structured outputs essential to our analysis.
\noindent
\textbf{(RQ1)} \changemarker{To evaluate inconsistencies across different sections of the responses}, we calculated the percentage\changemarker{s} of consistent and inconsistent samples as classified by GPT-4o. \changemarker{Although} LLMs generally produced consistent answer summaries, \changemarker{we observed significant inconsistencies} in other parts of the responses \changemarker{(see Table~\ref{tab:res_entailment})}. Consistent answer sections are \changemarker{mainly} related to individual considerations, public health, or professional advice. \textbf{(RQ2)} To pinpoint the disease categories where LLMs produced the most inconsistent answers, we analyzed the top three categories for each information unit with inconsistencies. \changemarker{LLMs} generated inconsistent answers across languages, about \changemarker{digestive systems, and endocrine and metabolic diseases}. Inconsistencies are \changemarker{notable} in an average of 71.97\% of answers referencing guidelines and evidence, and in 59.09\% of sentences within the latter disease category referencing health benefits and outcomes.
\begin{figure}[!b]
    \centering
    \begin{minipage}{0.45\textwidth}
        \centering
    \includegraphics[width=\linewidth]{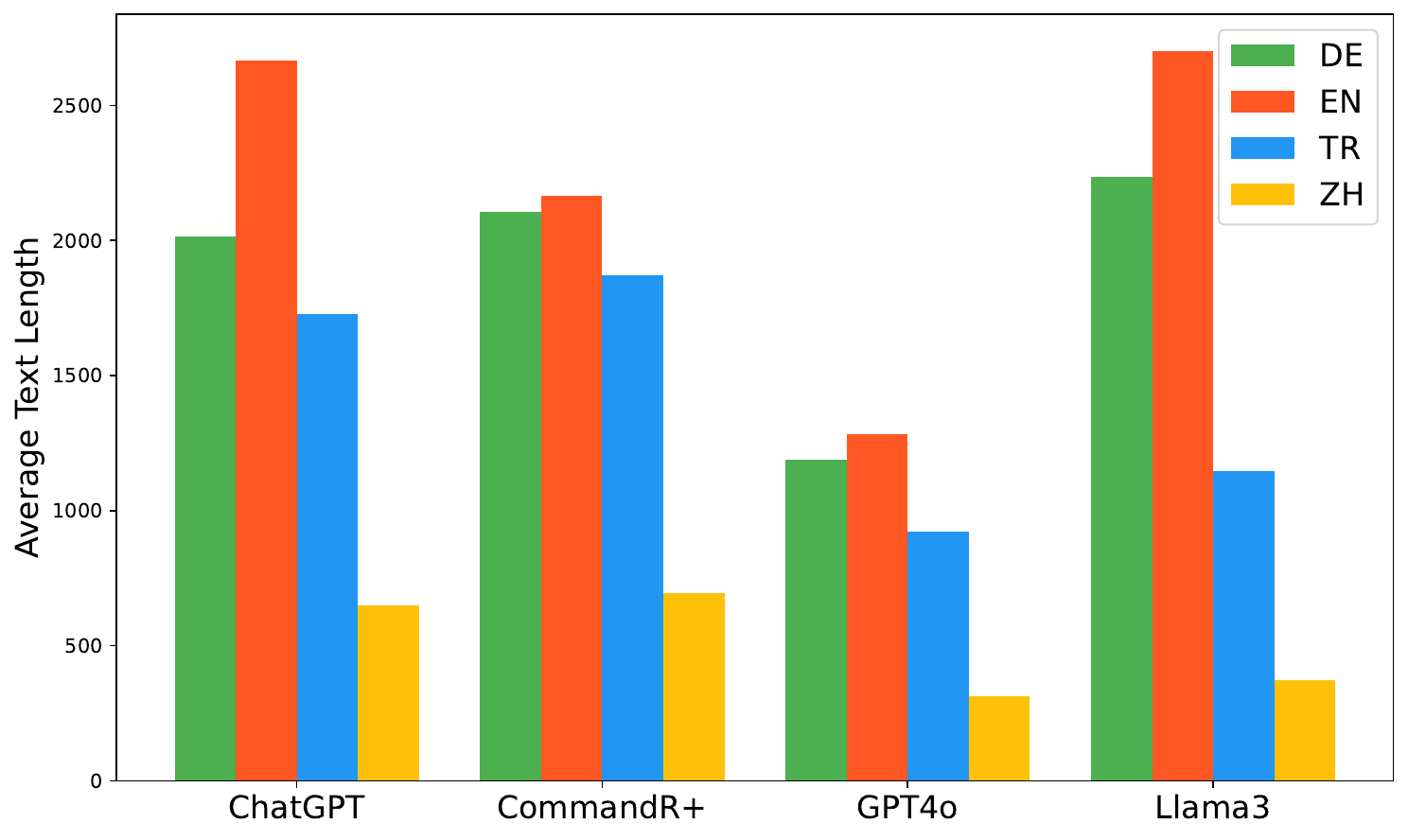}
    \caption{\changemarker{The LLMs tend to produce lengthy answers to the EN and DE questions.}}
    \label{fig:text_lens}
    \end{minipage}%
    \hspace{0.02\textwidth}  
    \begin{minipage}{0.45\textwidth}
        \centering
    \includegraphics[width=0.8\linewidth]{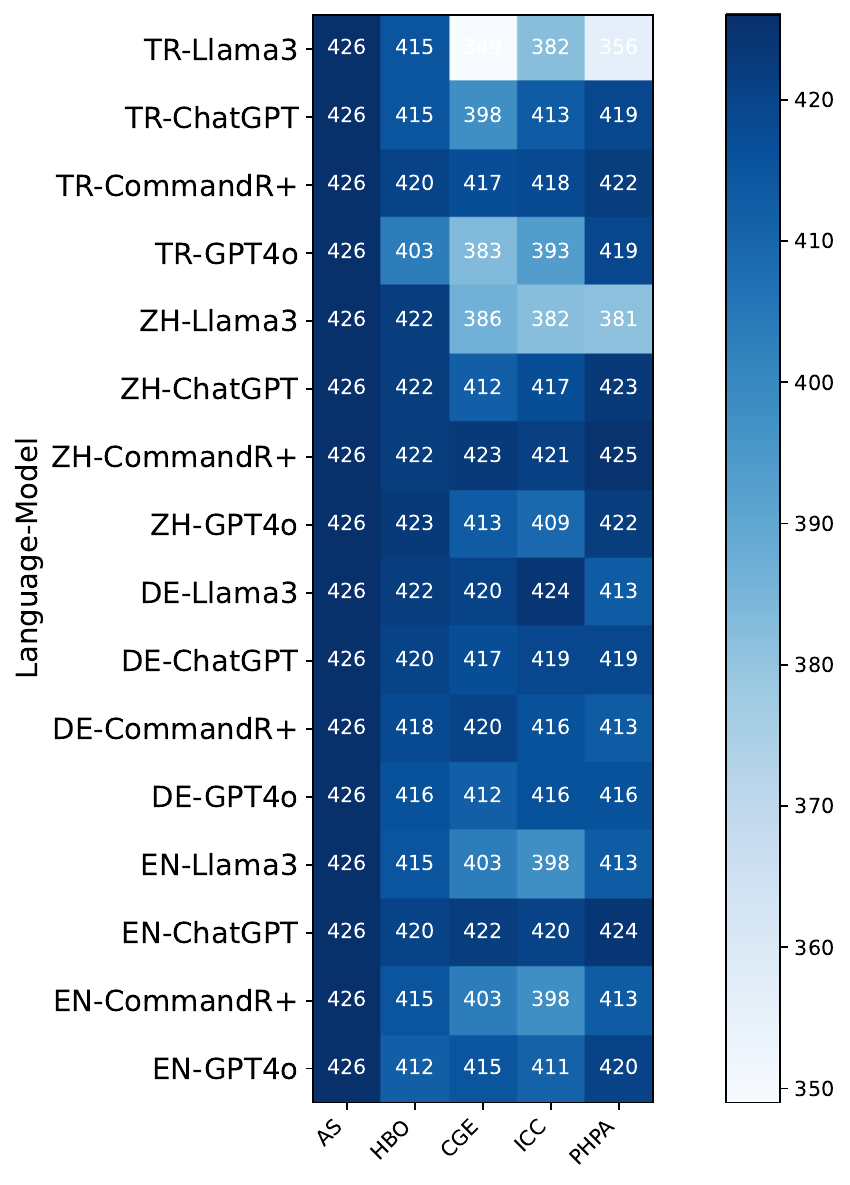}
    \caption{\changemarker{Answer occurence across the categories. TR answers omit some categories.}}
    \end{minipage}
\end{figure}

\begin{table}
\caption{Inconsistent Examples between the English answers and the answers in other languages \textcolor{blue}{\textit{with their English translations}}.} 
\centering
\includegraphics[trim={0.5cm 0cm 0.5 0cm},clip,width=\textwidth]{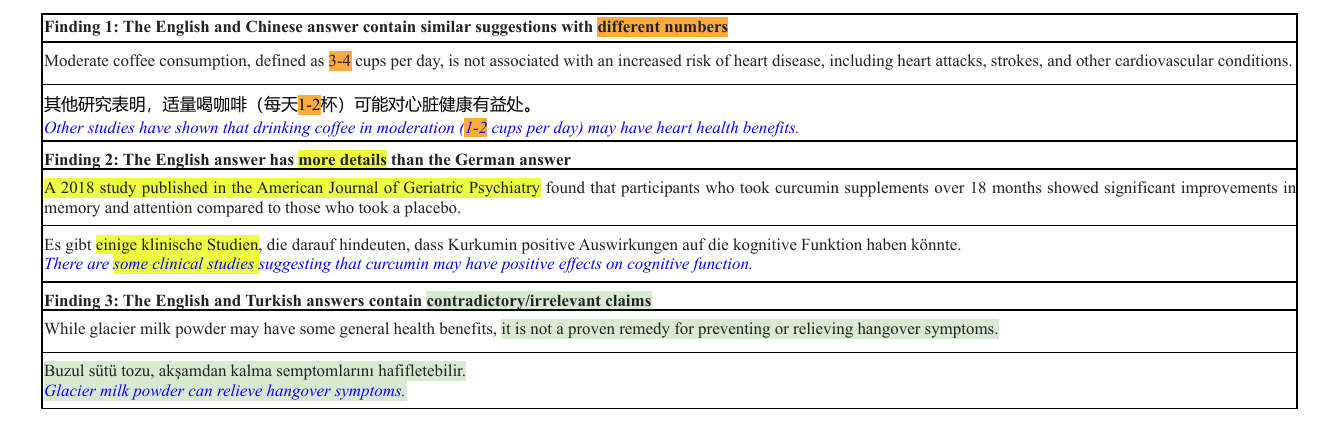}

    \label{fig:qualitative}
\end{table}

Finally, we performed a qualitative analysis on 50 samples per language used for human annotation. \changemarker{EN} responses often included more detailed information, such as references to research organizations and studies, specific findings, and precise examples, compared to other languages. Interestingly, this level of detail appeared even in cases deemed semantically consistent, which was not accounted for in our evaluation schema. Moreover, when answers included statistical or numerical details (e.g., research years or quantities), \changemarker{the values in EN responses often differed from those provided in other languages.} We also noted contradictory/irrelevant claims within a single response or between responses across different languages. \changemarker{Examples from the analysis are provided in Table~\ref{fig:qualitative}.}

\noindent
\changemarker{\textbf{Discussion.}}
\changemarker{As shown in Figure~\ref{fig:text_lens}, LLMs, particularly ChatGPT and Llama3, tend to generate longer answers in EN and DE. In contrast, TR answers often omit key elements like CGE, ICC, and PHPA, likely due to differences in training data and communication culture. For example, this may be affected by differences in the information available in different languages on Wikipedia~\cite{roy2022information} which is a common training source, or EN and DE often include more contextual detail due to their use in low-context cultures~\cite{hall1990understanding}. The closed models deliver more consistent answers, possibly due to regular updates and diverse human feedback.}
\\
\changemarker{\textbf{Limitations.}}
\changemarker{
The framework currently does not assess factual accuracy, as ground-truth data is only available in EN and DE. Creating such data requires medical expertise because treatments for some diseases (e.g., depression~\cite{skianis2024severity}) can vary across languages. In addition, parsing and labeling rely on GPT-4o, which may affect reproducibility. Despite human evaluation, errors in parsing and labeling could arise, particularly for information outside the current ontology. Furthermore, the single-question evaluation style restricts broader applicability, and GPT-4o struggles to detect fine-grained inconsistencies compared to human annotations. Finally, the analysis is limited to three non-EN languages, though the framework could be extended to others with expert support.}

\section{Conclusion} 
Our evaluation framework revealed notable differences between the answers in EN and other languages when addressing identical questions. These differences highlight the need for careful consideration of fairness implications when LLMs are deployed for medical applications, as they may affect equity in health communication and the potential for bias. \changemarker{In future work, we will explore open LLMs instead of GPT-4o to assess fine-grained inconsistencies and enhance the evaluation schema with detailed comparisons of information levels across languages.}

\begin{credits}
\subsubsection{\ackname}
\changemarker{The work of IBS and LF was supported by DynSoDA (funded by BMBF) and the Lamarr Institute for Machine Learning and Artificial Intelligence. The work of PR was supported by MCIN/AEI and by ERDF/EU (Grant No. PID2021-124361OB-C31). BS acknowledges support from the Horizon Europe project TANGO (Grant Agreement no. 101120763). Funded by the European Union. 
\subsubsection{\discintname}
Views and opinions expressed are however those of the author(s) only and do not reflect those of the funding organizations, EU or HaDEA.}
\end{credits}

\bibliographystyle{splncs04}
\bibliography{paper_bibliography}
\end{document}